\documentclass[letterpaper, 10 pt, conference]{ieeeconf}  

\IEEEoverridecommandlockouts                              

\usepackage{graphics} 
\usepackage{epsfig} 
\usepackage{here}
\title{\LARGE \bf
Imitation Learning for Object Manipulation
Based on Position/Force Information Using Bilateral Control
}

\author{Tsuyoshi~Adachi$^{1}$, Kazuki~Fujimoto$^{2}$, Sho Sakaino$^{3}$ and Toshiaki Tsuji $^{4}$
\thanks{$^{1}$Tsuyoshi Adachi is a student with the Department of Electronic Information System, School of Science and Technology, Saitama University,
        Saitama 338-8570, Japan.
        {\tt\small t.adachi.176@ms.saitama-u.ac.jp}}%
\thanks{$^{2}$Kazuki Fujimoto is a student with the Faculty of Engineering Department of Electrical and Electonic Engineering, Saitama University
        Saitama 338-8570, Japan.
        {\tt\small k.fujimoto.423@ms.saitama-u.ac.jp}}%
\thanks{$^{3}$S. Sakaino is with the Department of Electronic Information System, School of Science and Technology, Saitama University, 
        Saitama 338-8570, Japan , JST, PRESTO.
        {\tt\small sakaino@mail.saitama-u.ac.jp}}%
\thanks{$^{4}$T. Tsuji is with the Department of Electronic Information System, School of Science and Technology, Saitama University, 
        Saitama 338-8570, Japan.
        {\tt\small tsuji@mail.saitama-u.ac.jp}}%
}

\begin{document}

\maketitle
\thispagestyle{empty}
\pagestyle{empty}

\begin{abstract}
This study proposes an imitation learning method based on force and position information. Force information is required for precise object manipulation but is difficult to obtain because the acting and reaction forces cannot be separated. To separate the forces, we proposed to introduce bilateral control, in which the acting and reaction forces are divided using two robots. In the proposed method, two models of neural networks learn  a task; to draw a line along a ruler. We verify the possibility that force information is essential to imitate the human skill of object manipulation.

\end{abstract}

\section{Introduction}
Labor shortage is of concern in developed countries because of the declining population. Robots are expected to substitute human in doing simple factory tasks, but this type of work has yet to be robotized. One of the most difficult problems faced by current robots is their need to be able to adapt to the working conditions as robots are generally designed to only repeat specific tasks. Research to develop hardware to adapt to environmental changes has included flexible hands~\cite{hand} and suction hands~\cite{hand2}.
However, the physical characteristics of the hardware restricted the type of object that could be manipulated and it was difficult to operate objects that did not match the capability of the hardware~\cite{ex02}.

The adaptability of robots can be improved by means of software that is used to process large amounts of information. However, it is difficult to design the software and control systems because the behavior of a robot depends on much information. The idea of  "end to end learning" was proposed to reduce design efforts. In end to end learning, the behaviors of robots are determined only by sensor information, and agents are trained by machine learning. Levine~\it{et al. }\rm successfully manipulated objects by reinforcement learning using end to end learning over 800,000 trials using multiple manipulators~\cite{ex00}. 

However, even though multiple manipulators were prepared, it took a huge amount of time to learn the manipulations. Humans can easily adapt to perturbations in working conditions but considerable research into designing robots that can learn human manipulations is still required.
Technique have been actively conducted~\cite{ex04}~\cite{ex05}~\cite{ex06}. Recently, some studies have significantly reduced the number of trials by imitating human manipulative skills via remote control~\cite{ex01}. However, these conventional studies decided the behaviors of the robots based on position and image information but did not consider force information, which resulted in a low success rate. Some researches have been made to imitate human object manipulation techniques on robots based on force information~\cite{force_study1}. However, these studies do not use remote control system. On the contrary, in a peg-in-a-hole experiment using a remote control system, the feedback of the reaction force information to the operator improved the work efficiency \cite{ex03}. Reference \cite{ex03} implies that using force information in machine learning also improves the success rate in object manipulation. 
Previously, we also proved that every motion can be described as a combination of position and force controllers~\cite{sakaino}. 
 In order to use force information in machine learning, it is necessary to obtain the acting and reaction forces to mimic the human's force control. However, if a human directly manipulates and guides a robot to teach it a motion, the acting force and the reaction force cannot be separated because the forces are applied at the same place. Yokokura~\it{et al. }\rm demonstrated the use of bilateral control   to separate the forces in a behavior-cloning task~\cite{mc}.
\begin{figure*}[t]
\begin{center}
\includegraphics[width=150mm]{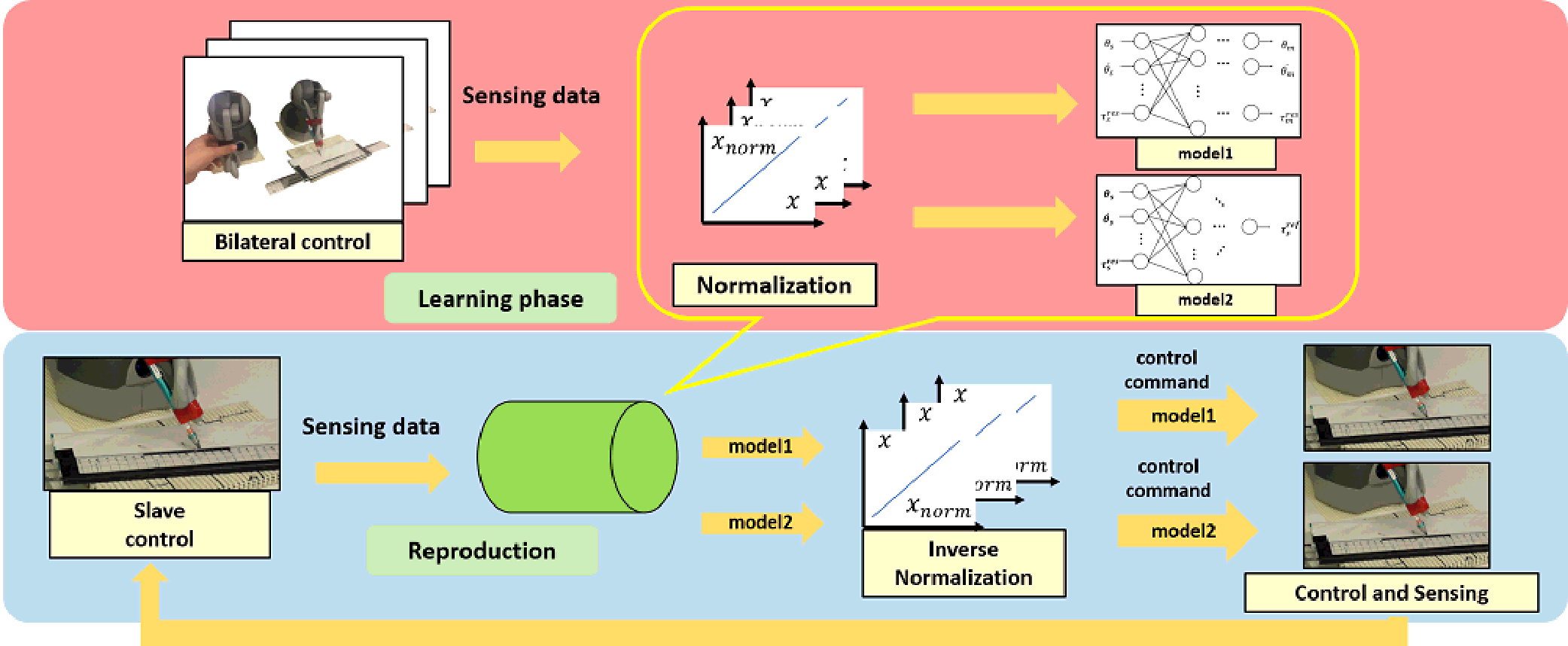}
\end{center}
\caption{Experiment flow}
\label{flow}

\end{figure*} 
Bilateral control is a remote control technology that uses two robots, one as a master robot and one as a slave robot~\cite{bilate1}~\cite{bilate2}. In this technology, an operator manipulates the master robot, and the slave robot tracks the master robot'~s position, while the reaction force of the slave robot is fed-back to the operator through the master robot. Then, the operator feels as if the operator is directly manipulating the remote environments of the slave side. The acting force is recoded only in the master robot while the reaction force is recorded only in the slave robot, resulting in a separation of the forces. Needless to say, position information may also be recorded by position and image sensors. 
Because the conventional method~\cite{mc} is just behavior-cloning, it has almost no adaptability to perturbations in the working conditions
Therefore, we propose neural networks to imitate human object manipulation skills using force and position information from motions in a variety of working conditions in order to achieve a standardized performance. As a result, an improvement in the success rate of object manipulation is expected. There is research on imitative learning by force-feedback-type bilateral control~\cite{force_study2}~\cite{force_study3}. However, there is no force controller in slave sides in force-feedback-type bilateral control. As a result, the imitated controllers do not include force control resulting in low adaptability to environmental perturbations. On the contrary, this research proposes to use 4-ch type bilateral control, which has position and force control both in master and slave side. A recurrent neural network (RNN) learns the relation between the input and output from human manipulation of an object by using the data obtained by bilateral control and then gives commands to the robots. In this paper, a robot conducted a task of drawing a line with a pencil along a ruler without using an image sensor but only using position and force information. Two models of RNNs were compared: a model that learns the position and force commands, and a model that learns the torque reference values. Machine experiments were carried out to investigate the success rate. The flow of the experiment in this paper is shown in Fig.~\ref{flow}.\\
The remainder of this paper is organized as follows. Section~II describes the manipulator used in this study. Section III describes the RNN network models and the normalization technique. Section IV describes the experimental results and compares the proposed models, and the work concludes with section V.

\section{Manipulator}
\begin{figure}[tbp]

\begin{center}
\includegraphics[width=50mm]{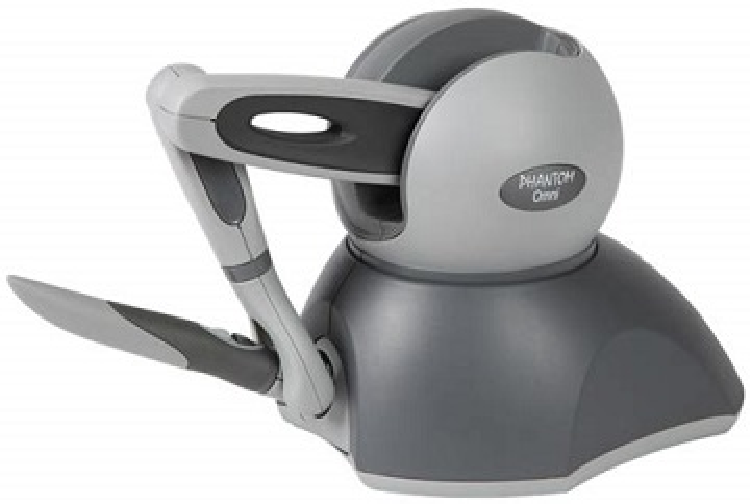}
\end{center}
\caption{Geomagic Touch}
\label{phantom}

\end{figure}

This section describes the manipulator used in this research, the Geomagic Touch manufactured by 3D Systems, shown in Fig.~\ref{phantom}.
Geomagic Touch was a 3-axis manipulator with a maximum demonstrating torque of 3.3 N. In this paper, we used two sets of Geomagic Touch for bilateral control. One of them was used as a master robot and the other was used as a slave robot.

\begin{figure}[tbp]
\begin{center}
\includegraphics[width=50mm]{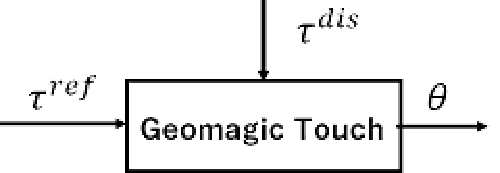}
\end{center}
\caption{Model of Geomagic Touch}
\label{phantom_model}

\end{figure}

\subsection{Control system }
\begin{figure}[tbp]

\begin{center}
\includegraphics[width=70mm]{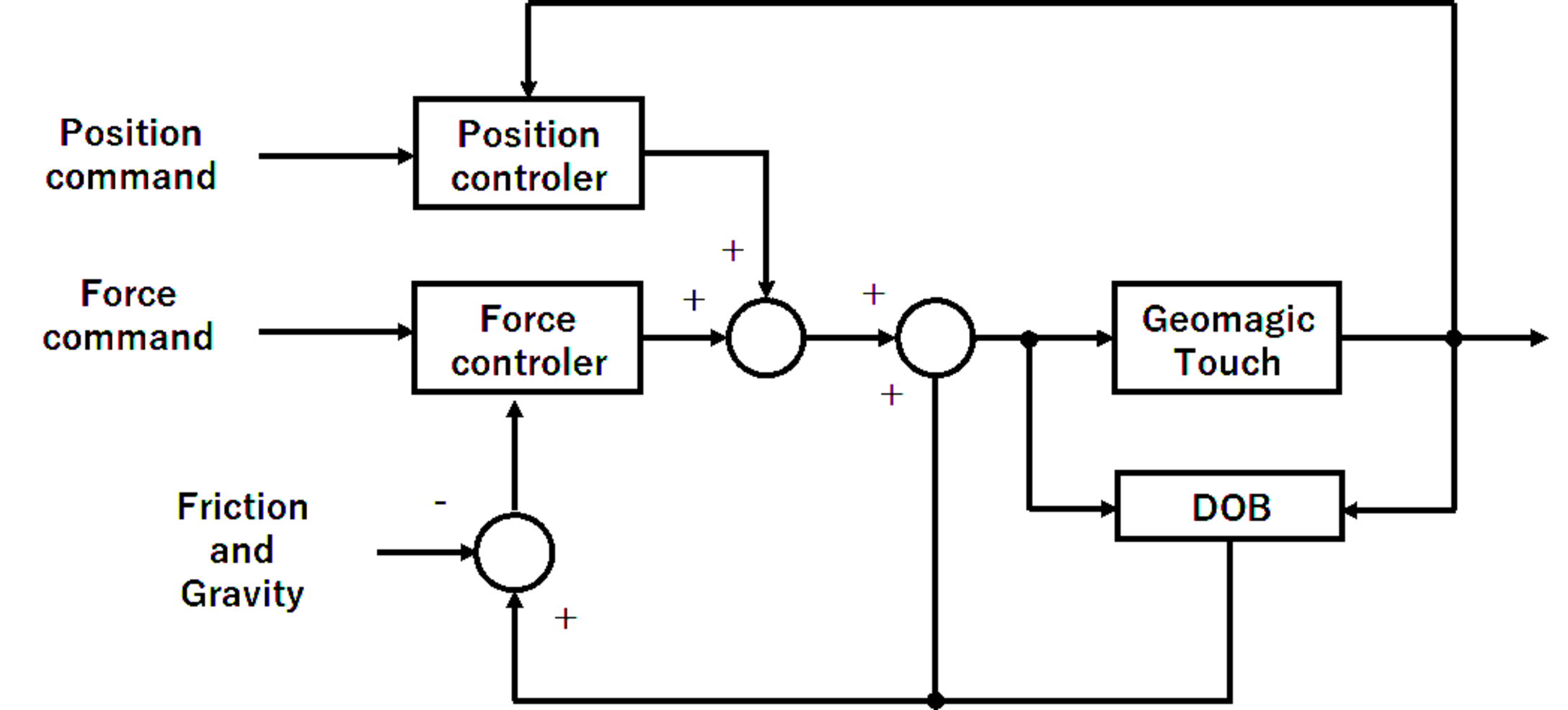}
\end{center}
\caption{Control system}
\label{control_system}
\end{figure}

\begin{figure}[tbp]
\begin{center}
\includegraphics[width=70mm]{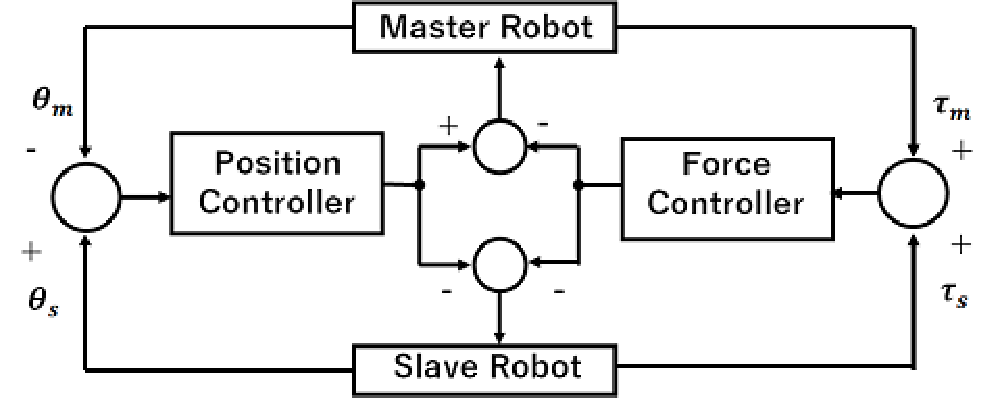}
\end{center}
\caption{4-ch bilateral controller}
\label{4ch}
\end{figure}

The Control parameters of Geomagic Touch were identified by using Yamazaki's method~\cite{id}. Fig.~\ref{phantom_model} shows a simplified model of Geomagic Touch; as the manipulator is affected by unexpected disturbances, a disturbance observer~\cite{DOB} was implemented to compensate for these disturbances. The block diagram of the control system of Geomagic Touch is shown in Fig.~\ref{control_system}. The controller is composed of a position controller and a force controller. The position controller gives a position control reference represented by (\ref{eq:p}). The force controller gives a force control reference represented by  (\ref{eq:q}).

\begin{eqnarray}
 \tau_{p} = (K_{p}+K_{v}s) (\theta^{cmd} - \theta^{res}) \label{eq:p}\\
 \tau_{f} = - K_{f}( \tau^{cmd} + \tau^{res}) \label{eq:q}
\end{eqnarray}
Here, $\theta$, $\tau$, $K_{p}$, $K_v$, and $K_{f}$ are angle, torque, position feedback gain, velocity feedback gain, and force feedback gain, respectively.
Since Geomagic Touch can only measure angle responses, each angular velocity and acceleration are derived by using pseudo derivatives. The reaction force was measured by using a reaction force observer~\cite{RFOB}. Bilateral control was implemented in two sets of Geomagic Touch as mentioned in the introduction. 
Bilateral control is a remote control technology, where the master robot 
operates the slave robot through a control system. In addition, when the slave robot receives a reaction force from the environments, force feedback is given to the master robot. We used a 4-ch bilateral controller~\cite{sakaino}, which is known to be the best bilateral controller. A block diagram of the 4-ch bilateral controller is shown in Fig.~\ref{4ch}. The 4-ch bilateral controller establishes the law of action and reaction by synchronization of the master and slave positions. Equations~(\ref{eq:9}) and~(\ref{eq:10}) shows the requirements.

\begin{eqnarray}
 \theta_{m} - \theta_{s} = 0 \label{eq:9}\\
 \tau_m^{res} + \tau_s^{res} = 0 \label{eq:10}
\end{eqnarray}
$\theta_{m}$ is the master angle. $\theta_{s}$ is the slave angle and $\tau_m{}^{res}$ is the reaction force of the master. $\tau_{s}^{res}$ is the reaction force of the slave.
Equations~(\ref{eq:bilate1}) and (\ref{eq:bilate2}) describe the input of the master and the slave with bilateral control where $J$ is the inertia. . 

\begin{eqnarray}
 \tau^{ref}_{s} = \frac{J}{2}(K_{p}+K_{v}s) (\theta^{res}_{m} - \theta^{res}_{s}) - \frac{K_{f}}{2}(\tau^{res}_{m} + \tau^{res}_{s})\label{eq:bilate1}\\
\tau^{ref}_{m} = \frac{J}{2}(K_{p}+K_{v}s) (\theta^{res}_{s} - \theta^{res}_{m}) - \frac{K_{f}}{2}(\tau^{res}_{m} + \tau^{res}_{s})\label{eq:bilate2}
\end{eqnarray}

\section{Imitation Learning}
\subsection{Recurrent neural network model}
A RNN is a network that holds the time series information. The network contributes to  natural language processing, voice processing, and the like~\cite{RNN1} \cite{RNN2}. Recently the network has been applied to manipulation of robots and was used in this research~\cite{RNN3}.

\begin{figure}[tbp]
\begin{center}
\includegraphics[width=45mm]{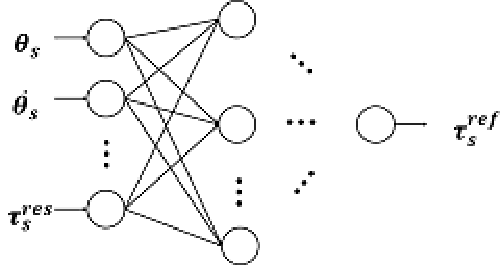}
\end{center}
\caption{Model to predict torque reference (model 1)}
\label{adachi_model}

\end{figure}

\begin{figure}[tbp]
\begin{center}
\includegraphics[width=45mm]{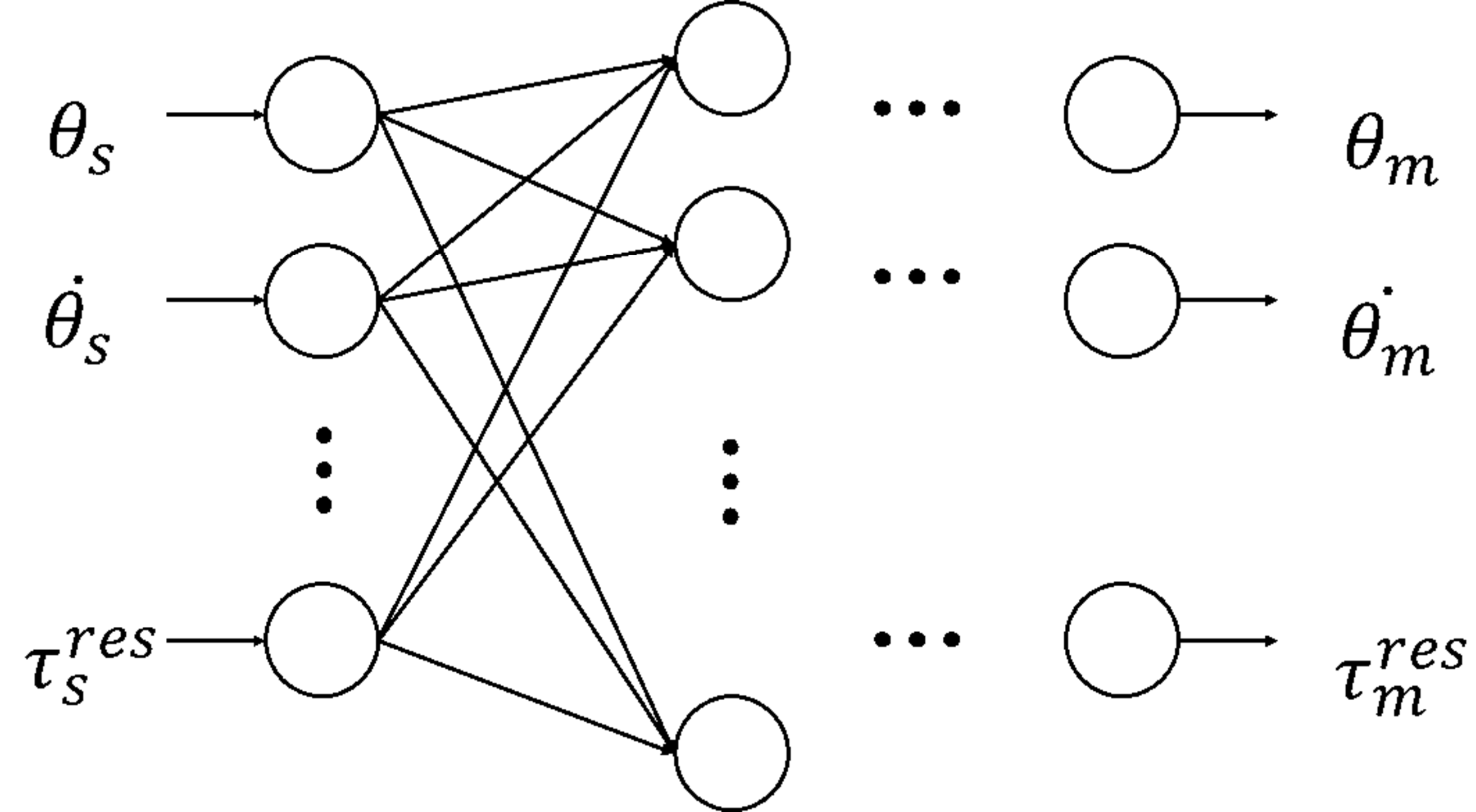}
\end{center}
\caption{Model to predict angle, angular velocity, and torque commands (model 2)}
\label{fujimoto_model}

\end{figure}

Figs.~\ref{adachi_model} and~\ref{fujimoto_model} show the proposed RNNs using position (angle) and force (torque) information obtained by bilateral control. The model in Fig.~\ref{adachi_model} predicts the next torque references of the slave robot by using angle, angular velocity, and torque responses~(model 1). In order to train the model, the angle, angular velocity, torque responses, and the torque references were collected by bilateral control. That is, the network model learns not only the position and force commands, but also the controller itself. On the other hand, the model shown in Fig.~\ref{fujimoto_model} predicts the angle, angular velocity, and torque commands by using the angle, angular velocity, and torque responses~(model2). This model only predicts the commands and the position and the force controllers are the same as those described in Fig.~\ref{4ch}. Because of the calculation time of the RNN, there was a delay between a RNN program written by python using Chainer, and a control program written by C Language. The delay was less than 20 msec, and therefore, the RNNs predicted data 20 msec in the future.


\subsection{Normalization}

Normalization is a technique to normalize input data and output data to equalize a data range.  In this research, because the angle, angular velocity, and torque have quite different ranges they must be normalized. The normalization function is shown in (\ref{eq:11}).
\begin{equation}
  x_{norm} = \frac{x - x_{min}}{x_{max}-x_{min}} \label{eq:11}
\end{equation}

Here, $x_{norm}$ is the new training data after normalization. $x_{min}$ and $x_{max}$ are the maximum and minimum values of the movable range. Since the outputs of the RNN are the normalized values, the outputs are denormalized in calculating the actual reference or command values.
\section{experiment}

\subsection{Parameter}

Gravity and friction force compensation were incorporated into the control system based on the parameters in Table~\ref{tb:parameter}.  $1, 2, 3$ are the number of robot's joints.

\begin{table}[htb]
  \begin{center}
    \caption{Control Parameter}
    \begin{tabular}{|c|r|c|} \hline
      parameter &  coefficient & unit \\ \hline
      $g$ & 40& rad/s \\ \hline
      $J_1$ & 4.0& ${\rm mkgm^2}$\\ \hline
      $J_2$ & 8.21& ${\rm mkgm^2}$   \\ \hline
      $J_3$ & 3.43& ${\rm mkgm^2} $  \\ \hline
      $M_2$ & 95& mNm\\ \hline
      $M_3$ & 95& mNm \\ \hline
      $D$ & 12 & Nms/rad\\ \hline
      $K_{p}$ & 100 & \\ \hline
      $K_{v}$ & 20& \\ \hline
      $K_{f}$ & 1.0 & \\ \hline
    \end{tabular}
    \label{tb:parameter}
  \end{center}
\end{table}

\subsection{Training data set}

\begin{figure}[tbp]
\begin{center}
\includegraphics[width=50mm]{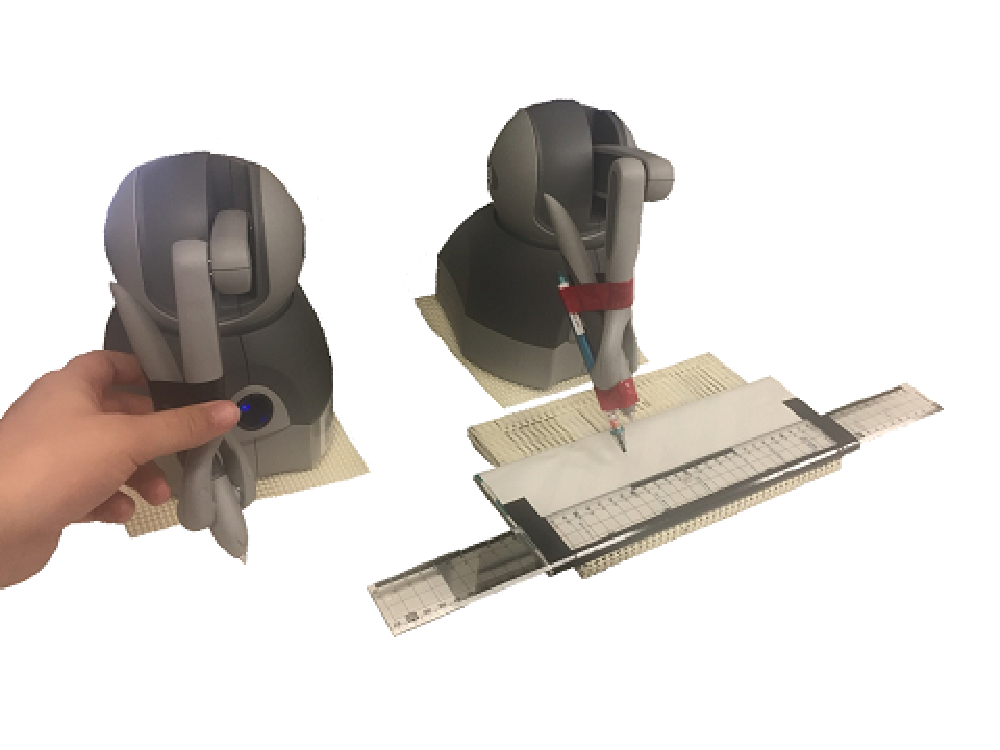}
\end{center}
\caption{Situation of bilateral control}
\label{control_situation}
\end{figure}

\begin{figure}[tbp]
\begin{center}
\includegraphics[width=70mm]{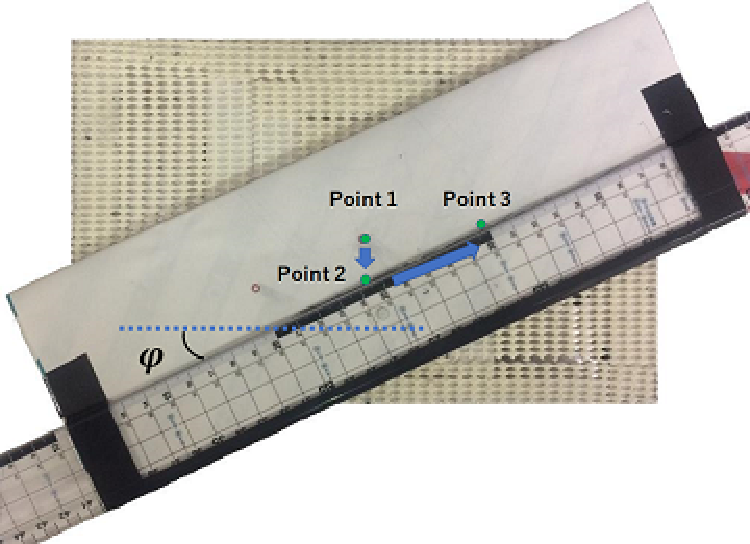}
\end{center}
\caption{Definition of points and inclination}
\label{fig:slope}
\end{figure}

Acquisition of learning data is an important phase in machine learning. As explained in section I\hspace{-.1em}I, training data was acquired from the robots using bilateral control. The master robot was operated directly by an operator while a pencil was fixed to the slave robot. The situation of bilateral control is shown in Fig.~\ref{control_situation}. The fixed pencil did not reach the floor because there was a limit in the working space of Geomagic Touch. Therefore, by placing a pedestal on the floor of the slave side, the pencil was brought into contact with the paper surface. In order to draw a line using a ruler, the ruler was fixed to the pedestal. The state where the ruler was parallel to the 
pedestal was set to 0 degrees and the three inclination states of 0 degrees, 30 degrees and 60 degrees were to draw the line. Then, the master and the slave's angle, angular velocity, torque response, and torque reference values were saved. The definition of 
the inclination is shown in Fig.~\ref{fig:slope}. The method for acquiring the training data comprised from of three steps, which are detailed below. 

\begin{enumerate}
  \item Step1\\
  Fit the initial position of the manipulator to point 1  (at this time, since the initial data was the synchronization data of master and slave position, we discarded the data from the first 5 seconds.)
  \item Step2\\
  Draw a line from point 1 to point 2
  \item Step3\\
  Draw a line from point 2 to point 3 along the  ruler after the pencil has contacted the ruler (the duration from Step 1 to Step 3 was less than 5 seconds).
\end{enumerate}

\begin{figure}[htbp]
 \begin{minipage}{0.3\hsize}
  \begin{center}
   \includegraphics[width=30mm]{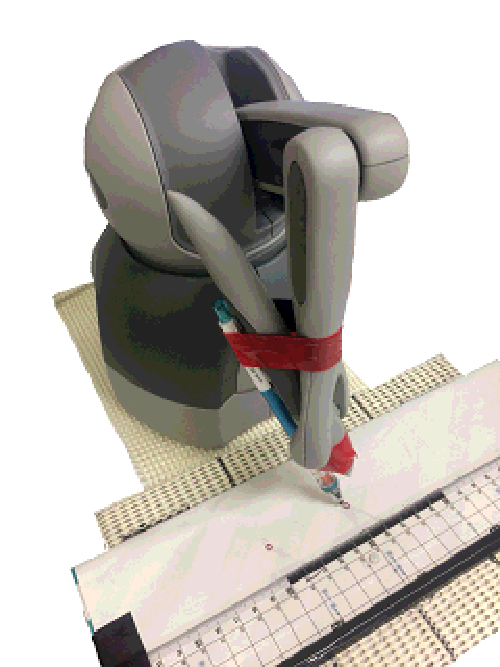}
  \end{center}
  \caption{Experiment (Step1)}
  \label{fig:one}
 \end{minipage}
 \begin{minipage}{0.3\hsize}
 \begin{center}
  \includegraphics[width=30mm]{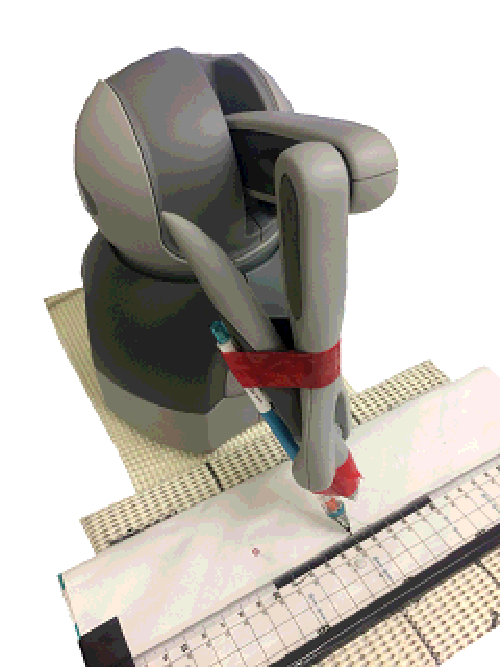}
 \end{center}
  \caption{Experiment (Step2)}
  \label{fig:two}
 \end{minipage}
 \begin{minipage}{0.3\hsize}
 \begin{center}
  \includegraphics[width=30mm]{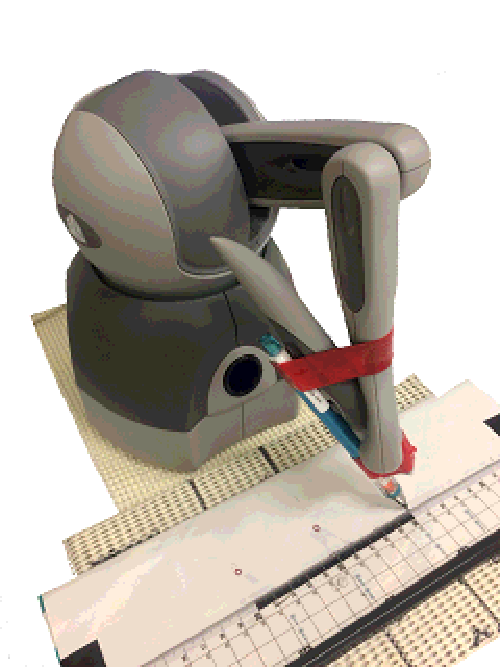}
 \end{center}
  \caption{Experiment (Step3)}
  \label{fig:steps}
 \end{minipage}
\end{figure}
 In order to normalize the saved data, the maximum and minimum values of the data were set. The setting of the maximum value and the minimum value of the movable area are shown in Tables II and III. Then number 1, 2, and 3 indicate the lowest joint, the middle joint, and the top joint, respectively.


\begin{table}[htb]
\begin{center}
\caption{model~1 parameter}
    \begin{tabular}{|c|r|r|} \hline
      parameter &max & min \\ \hline
      $\theta_{1s}$ & 0.5 rad & -0.5 rad\\ \hline
      $\theta_{2s}$ & 0.4 rad & 0.1 rad\\ \hline
      $\theta_{3s}$ & 0.5 rad & 0.1 rad   \\ \hline
      $\dot\theta_{1s}$ & 0.05 rad/s & -0.35 rad/s   \\ \hline
      $\dot\theta_{2s}$ & 0.05 rad/s & -0.20 rad/s   \\ \hline
      $\dot\theta_{3s}$ & 0.35 rad/s & -0.05 rad/s   \\ \hline
      $T_{1res}$ & 50 mNm & -250 mNm  \\ \hline
      $T_{2res}$ & 50 mNm & -600 mNm  \\ \hline
      $T_{3res}$ & 100 mNm & -100 mNm  \\ \hline
      $T_{1ref}$ &  20 mNm & -20 mNm  \\ \hline
      $T_{2ref}$ &  25 mNm & -25 mNm  \\ \hline
      $T_{3ref}$ &  15 mNm & -15 mNm  \\ \hline
\end{tabular}
\end{center}
\label{tb:norm1}
\end{table}

\begin{table}[htb]
\begin{center}
\caption{model2 parameter}
\begin{tabular}{|c|r|r|}\hline
      parameter &max & min \\ \hline
      $\theta_{1m}$ & 0.5 rad & -0.5 rad\\ \hline
      $\theta_{2m}$ & 0.4 rad & 0.1 rad\\ \hline
      $\theta_{3m}$ & 0.5 rad & 0.1 rad   \\ \hline
      $\dot\theta_{1m}$ & 0.05 rad/s & -0.35 rad/s   \\ \hline
      $\dot\theta_{2m}$ & 0.05 rad/s & -0.20 rad/s   \\ \hline
      $\dot\theta_{3m}$ & 0.35 rad/s & -0.05 rad/s   \\ \hline
      $T_{1res}$ & 250  mNm & -250 mNm \\ \hline
      $T_{2res}$ & 600  mNm & -600 mNm \\ \hline
      $T_{3res}$ & 100  mNm & -100 mNm\\ \hline

\end{tabular}
\end{center}
\label{tb:norm2}
\end{table}

\subsection{RNN learning}
In order to compare the two models, the RNNs are trained using the same motion but  the output data differed. The composition of each RNN is shown in Table \ref{tb:compose}. Fourth layer was the all connected layer. An activation function of the long-short term memory (LSTM) was a hyperbolic tangent. In the training, one of 15 pieces of training data was randomly selected. Then, two seconds of data were randomly extracted from the time-series-data for further training.


\begin{table}[htb]
  \begin{center}
    \caption{recurrent neural network component}
    \begin{tabular}{|c|c|c|c|c|} \hline
      &first layer & second layer & third layer & fourth layer\\ \hline
      units & input data &100 & 100 & output data\\ \hline
      LSTM & ~& Z & Z&~\\ \hline
    \end{tabular}
    \label{tb:compose}
  \end{center}
\end{table}

\subsection{Experiment}

\begin{figure}[htbp]
 \begin{minipage}{0.3\hsize}
  \begin{center}
   \includegraphics[width=25mm]{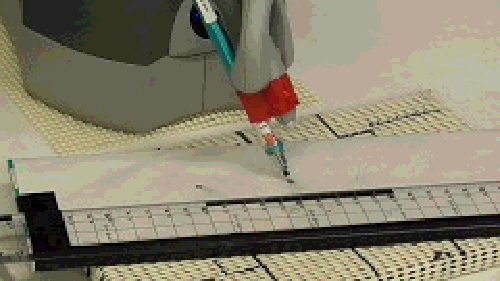}
  \end{center}
  \caption{Snap shot 1}
  \label{fig:snap1}
 \end{minipage}
 \begin{minipage}{0.3\hsize}
 \begin{center}
  \includegraphics[width=25mm]{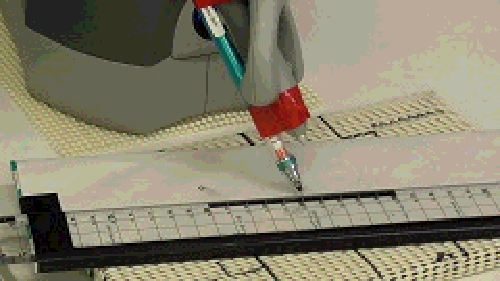}
 \end{center}
  \caption{Snap shot 2}
  \label{fig:snap2}
 \end{minipage}
 \begin{minipage}{0.3\hsize}
 \begin{center}
  \includegraphics[width=25mm]{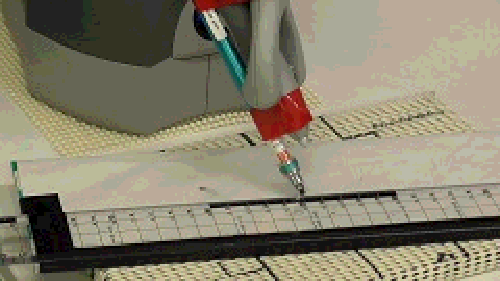}
 \end{center}
  \caption{Snap shot 3}
  \label{fig:snap3}
 \end{minipage}

 \begin{minipage}{0.3\hsize}
  \begin{center}
   \includegraphics[width=25mm]{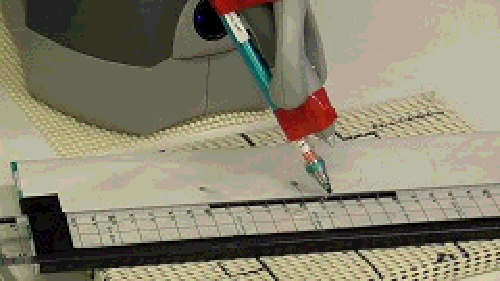}
  \end{center}
  \caption{Snap shot 4}
  \label{fig:snap4}
 \end{minipage}
 \begin{minipage}{0.3\hsize}
 \begin{center}
  \includegraphics[width=25mm]{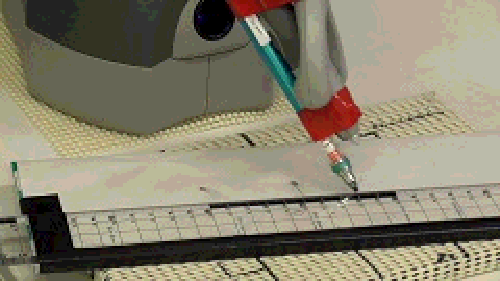}
 \end{center}
  \caption{Snap shot 5}
  \label{fig:snap5}
 \end{minipage}
\end{figure}

The performance of the RNN models were experimentally evaluated by drawing a line along the ruler with the inclination of the ruler at 15 degrees and 45 degrees, inclinations which were not included in the training data.  The task was regarded as a success if the robot drew lines 2 cm or longer along the ruler. 
Figs.~\ref{fig:snap1},~\ref{fig:snap2},~\ref{fig:snap3},~\ref{fig:snap4}, and~\ref{fig:snap5} show snap shots of the successful drawing of a line along the ruler.

In each model, the success and failure rates are shown in Table~\ref{tb:success_rate}.

\begin{table}[htb]
  \begin{center}
    \caption{machine experiment result}
    \begin{tabular}{|c|r|c|} \hline
      model& inclination & success rate\\ \hline
      $1$ & 15 degree & 90\% \\ \hline
      $2$ & 15 degree & 75\%  \\ \hline
      $1$ & 45 degree &  65\% \\ \hline
      $2$ & 45 degree & 70\% \\ \hline
    \end{tabular}
    \label{tb:success_rate}
  \end{center}
\end{table}

At an inclination of 15 degrees, model 1 had a better success rate than model 2, while the success rate at 45 degrees was almost the same between both models. 
Therefore, model 1 seems to be better than model 2.
 However, model 2 was more stable. 
Here, the differences in the failures of each model are discussed in further detail. 

The situation of failure in model 1 differed with the inclination. At an inclination of 15 degrees, the manipulator applied a force in the direction toward the ruler, and could not move. On the contrary, at an inclination of 45 degrees, the torque reference tended to diverge and became unstable. 
In the failures of model 2, there were a few dependencies on inclination. The pencil could not keep contact with the paper regardless of the inclination. Note that even though the task was not accomplished, the robot was stable. 
Since model 1 learned not only the position and force commands but also the controllers, the stability of the control was quite difficult to guarantee. However, because model 2 learned only the position and force commands and a stable controller was designed by using control theories, the robot moved to inappropriate commands even when it failed. 
Therefore, we believe model 2 has more potential to obtain a better performance. For example, if image information is also utilized for imitation learning, the success rate will be improved.

\begin{figure}[H]
\begin{center}
\includegraphics[width=60mm]{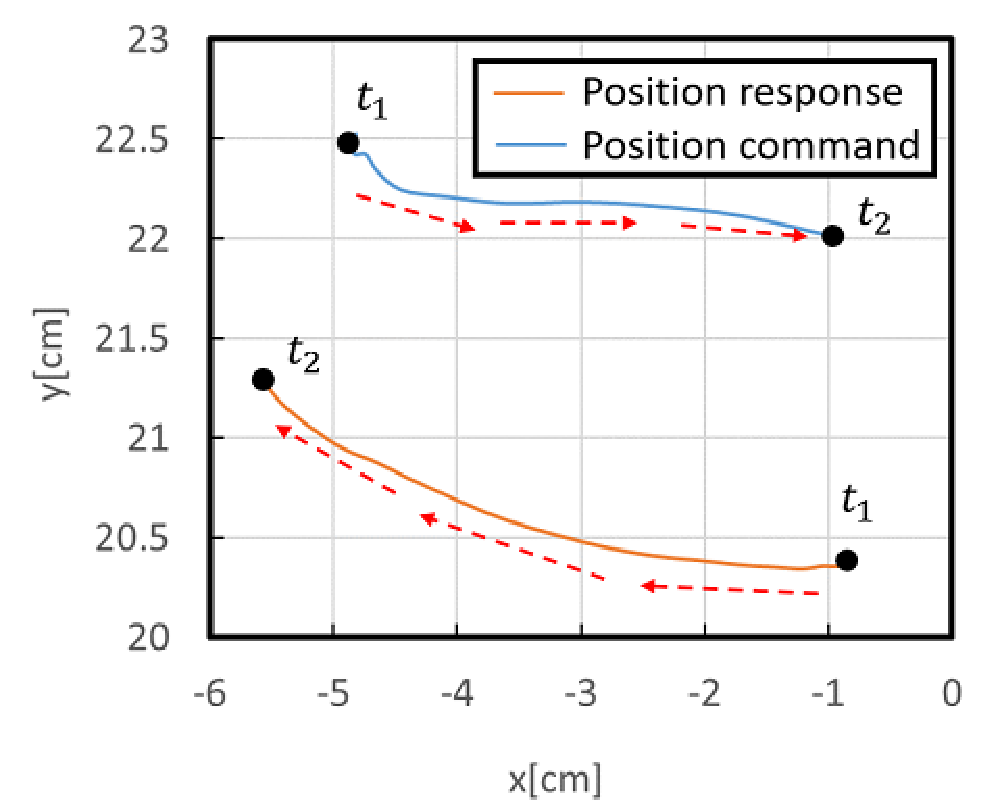}
\end{center}
\caption{Drawing with proctractor and position command predicted by model2}
\label{fig:success}
\end{figure}

\begin{figure}[H]
\begin{center}
\includegraphics[width=60mm]{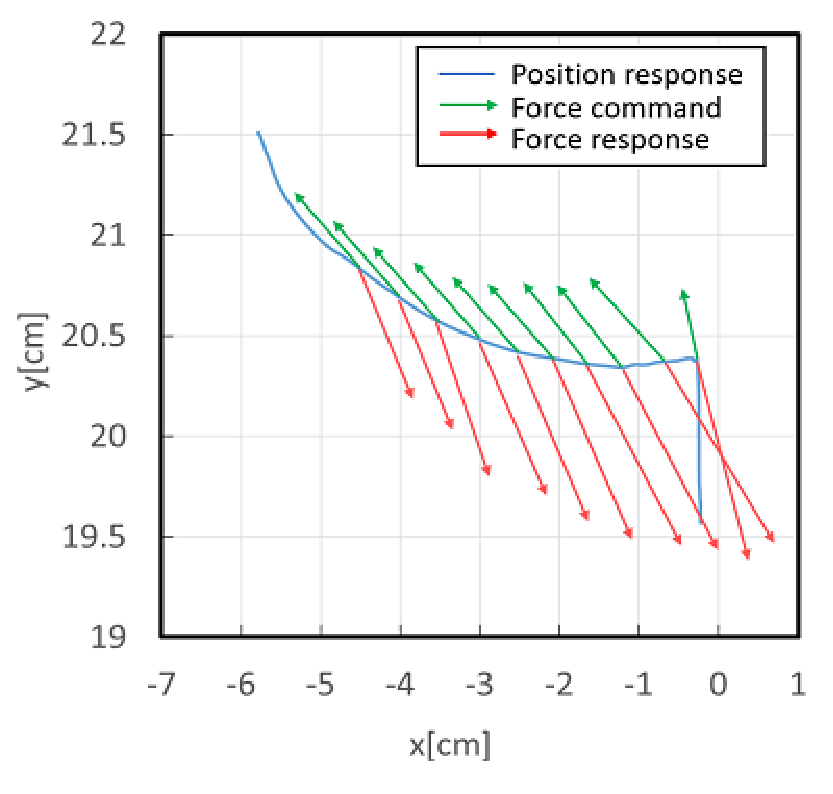}
\end{center}
\caption{Response and command forces drawing a line along the proctractor (model~2)}
\label{fig:vector2}
\end{figure}

To demonstrate further the performance of the set-up, a task to draw a curve by using a protractor was conducted using model 2. Fig.~\ref{fig:success} shows that the robot could draw a curve using model 2 even though there was no training data for the use of a protractor. Fig.~\ref{fig:vector2} shows the trajectory of the position response  and the vectors of the force command and response when a curve was drawn along the protractor. 

\begin{figure}[H]
\begin{center}
\includegraphics[width=45mm, angle = 90]{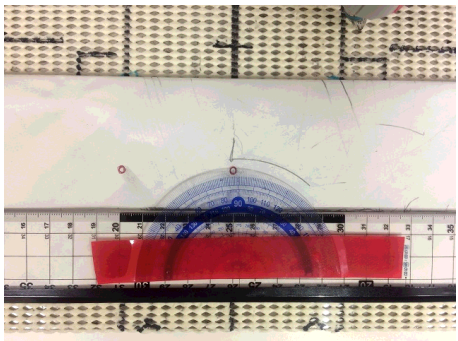}
\end{center}
\caption{Photograph of a trajectory drawing with proctractor}
\label{fig:proctractor}
\end{figure}

 As can be seen from Figs.~\ref{fig:success} and~\ref{fig:vector2}, model 2 predicted the position command value inside the protractor and the force command value along the protractor. This result shows the force controller made a trajectory along the protractor and  the position controller made pressing force against the protractor. We separated the work of drawing a curve using 4-ch bilateral control into position and force information, then the RNN learned imitating work of human drawing a curve from sensor information. Fig.~\ref{fig:vector2} is the trajectory of the pen derived from the sensors. Fig.~\ref{fig:proctractor} is a photograph of a curve drawn along the protractor. From Figs.~\ref{fig:vector2} and~\ref{fig:proctractor}, the trajectory read from the sensors was not an accurate arc. In other words, the sensors of the manipulator used in this experiment did not have high accuracy. However, the arc actually drawn was accurate.  Therefore it can be said that the human object manipulation technique was reproduced without using high-performance position sensors or controller but force controller.


\section{Conclusion}
In this paper, we proposed two RNN models that imitated object manipulation by humans using both position and force information with the help of bilateral control. To compare the two models, a task to draw a line along a ruler was given to the robot systems, and a robot succeeded to draw lines even with untrained inclination. Furthermore, the robot could draw a curve by using a protractor without any preliminary knowledge of the protractor. This adaptability in object manipulation was obtained by force control. Usually, image information has been exploited to adapt to changes in environments. However, it is quite difficult to detect the contact state only by image information, but with force control it is possible. This paper 
demonstrated the importance of force information in the machine learning of object manipulation. Also, bilateral control is a key technique to obtain training data including force information. This study is a simple step in the progress toward general object manipulation. However, more complicated tasks will be realized by integrating image information and increasing the number of joints. When it conducts not learning task, model 2 can control stably.


\setlength\floatsep{2pt}

\setlength\floatsep{2pt}

\addtolength{\textheight}{-12cm}   




\section*{ACKNOWLEDGMENT}

This work was supported by JST, PRESTO Grant Number JPMJPR1755, Japan.


\end{document}